%% file: main.tex
\documentclass[sigconf]{aamas} 

\usepackage{hyperref}       
\usepackage{balance}
\usepackage[utf8]{inputenc} 
\usepackage[T1]{fontenc}    
\usepackage{url}            
\usepackage{booktabs}       
\usepackage{amsmath} 
\usepackage{amsfonts}       
\usepackage{nicefrac}       
\usepackage{microtype}      
\usepackage{xcolor}         
\usepackage{ulem}
\usepackage{graphicx} 
\usepackage{algorithm} 
\usepackage{algorithmic} 
\usepackage{multicol}

\title{Emergency action termination for immediate reaction in hierarchical reinforcement learning}

\author{Michał Bortkiewicz$^1$, Jakub Łyskawa$^1$, Paweł Wawrzyński$^1,2$, Mateusz Ostaszewski$^1$, Artur Grudkowski$^1$ and Tomasz Trzciński$^1,3,4,5$}

\affiliation{
  \institution{$^1$Warsaw University of Technology, Institute of Computer Science, $^2$Ensavid, \and $^3$Jagiellonian University, $^4$Tooploox, $^5$IDEAS NCBR}
  \city{}
  \country{}
  }

\begin{abstract}
Hierarchical decomposition of control is unavoidable in large dynamical systems. In reinforcement learning (RL), it is usually solved with subgoals defined at higher policy levels and achieved at lower policy levels. Reaching these goals can take a substantial amount of time, during which it is not verified whether they are still worth pursuing. However, due to the randomness of the environment, these goals may become obsolete. In this paper, we address this gap in the state-of-the-art approaches and propose a method in which the validity of higher-level actions (thus lower-level goals) is constantly verified at the higher level. If the actions, i.e. lower level goals, become inadequate, they are replaced by more appropriate ones. This way we combine the advantages of hierarchical RL, which is fast training, and flat RL, which is immediate reactivity. We study our approach experimentally on seven benchmark environments.
\end{abstract}


\input{macros}
\newcommand\wyciete[1]{}

\begin{document}

\pagestyle{fancy}
\fancyhead{}


\maketitle


\section{Introduction}



It has been shown that hierarchical reinforcement learning performs remarkably well on many complex tasks~\cite{gehring2021hierarchical,gurtler2021hierarchical,nachum2018data,levy2017learning,ghosh2019learning,eysenbach2019search}. This is due to the fact the control or sequential decision making in complex dynamical systems is often easier to synthesize when decomposed hierarchically~\cite{nachum2019hierarchy}. The high-level agent breaks down the problem into a series of sub-goals to be sequentially executed by the low-level agent.

\begin{figure}
    \centering
    \includegraphics[width=.3\linewidth]{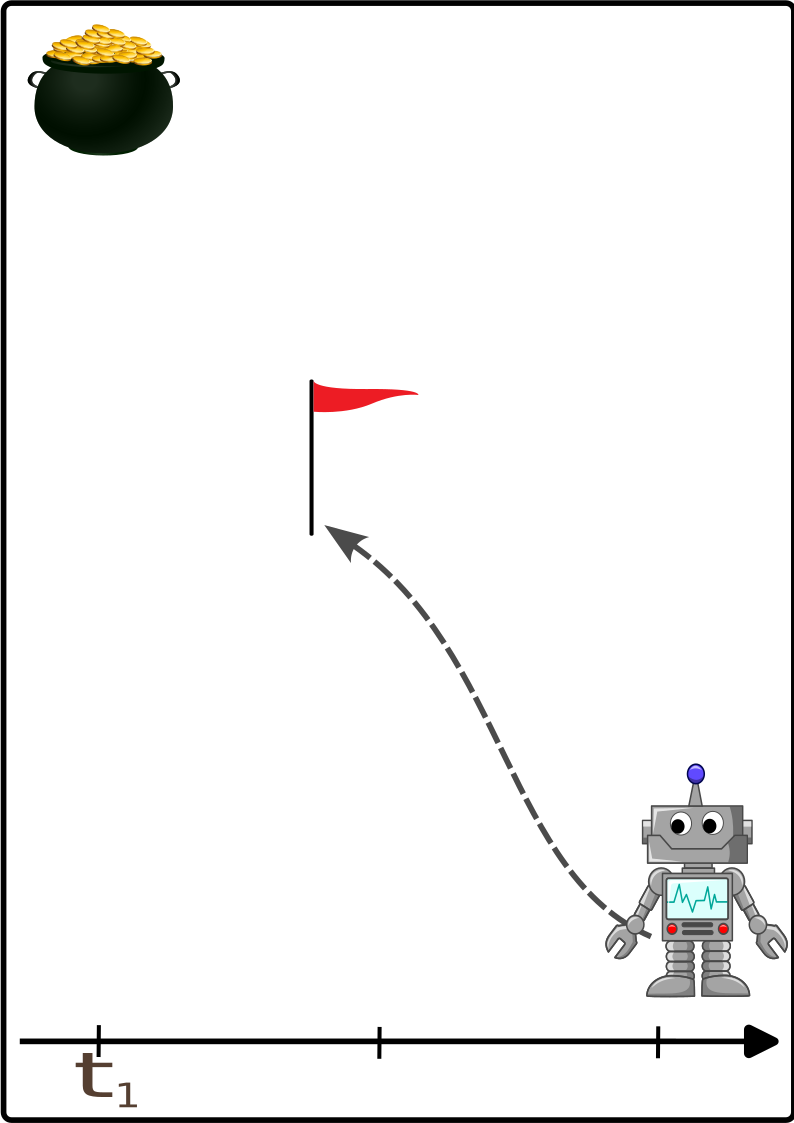}
    \includegraphics[width=.3\linewidth]{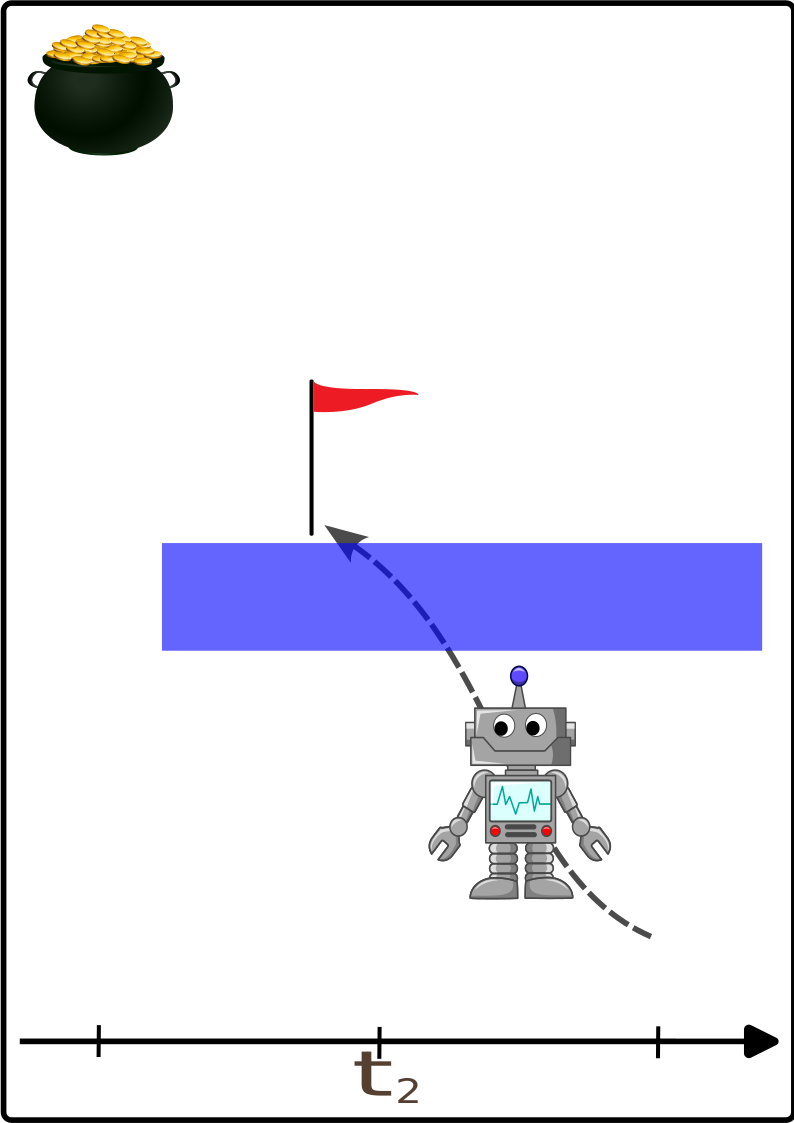}
    \includegraphics[width=.3\linewidth]{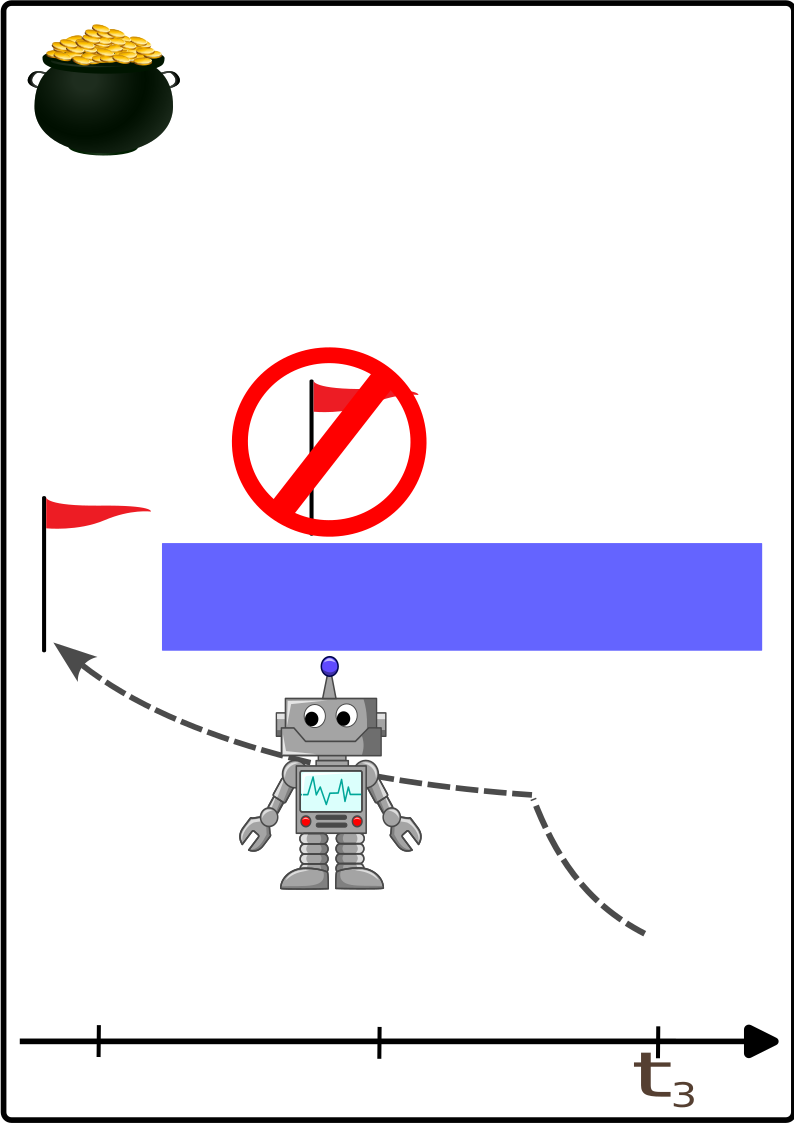}
    \caption{Emergency action termination in hierarchical RL: {\it Left:} Pursuing the long-term goal, the agent defines a~short-term one. {\it Middle:} When reaching the short-term goal, the agent finds out that this goal is obsolete due to a~random change in the environment. {\it Right:} Immediately, the agent defines a~different short-term goal. }
    \label{fig:teaser}
\end{figure}

However, the true potential of hierarchical learning methods has not been fully explored.
Many current approaches in hierarchical reinforcement learning has been designed by assuming environmental determinism. However, in the real world, this assumption is not true, since phenomena occur irregularly. In this paper, we propose a solution that is capable of operating in environments with random dynamics.

So far, a strong emphasis has been placed on solving problems arising from the complex dynamics of the interaction between agents at different levels~\cite{levy2017learning,nachum2018data, gurtler2021hierarchical}. However, the proposed solutions are not designed to operate in randomized dynamic environments -- dynamics in which some events occur in an irregular and therefore unpredictable manner.
  Many solutions assume that a high-level policy designates the sub-goal for a fixed period of time. During this time, the low-level agent is expected to complete the sub-goal.
  The duration of such a high-level action is either set by a human expert as a hyperparameter~\cite{mcgovern2001automatic,vezhnevets2017feudal,levy2017learning,nachum2018data} or it is a part of the output of a high-level agent~\cite{gurtler2021hierarchical}. 
  This duration generally does not change in response to random events. If during the pursuit of the sub-goal by a low-level agent, any circumstances arise that make this plan obsolete, then we have to wait until the end of the sub-goal anyway. In such a situation either an inadequate behavior is exercised or we hope that the low-level agent will set a new plan for itself thereby taking over the role of the higher-level agent.




 How the current solutions of the hierarchical reinforcement learning are not adapted to the dynamic environments was already shown by~\citet{gurtler2021hierarchical}. The authors demonstrate that in the environments requiring precise timing of low and high level policies, it is important to adapt the duration of high-level action. Therefore, in~\cite{gurtler2021hierarchical}, the high-level agent returns the sub-goal and the time in which it is to be achieved as parts of its action. This precise communication is effective in dynamic situations in which meticulously pursued sub-goals allow for better planning.

In this paper, we are going one step further and consider the problem of dynamic environments in which the agent is unable to accurately predict the future. Consequently, it is essential that a high-level policy is able to react immediately to random situations and to replace current sub-goals. 

 Suppose the state coordinates not being under direct agent control are subject to some random dynamics. 
 In particular, we can imagine a situation where the agent is a ship and its aim is to cross a drawbridge, which opens in an irregular manner, e.g. due to varying traffic volumes.
 Since we are unable to predict such a situation, we need to react to it. 
Such unexpected events can be anticipated in two ways in the planning process. One can designate short-range sub-goals so that each is determined in relation to the most current state of the environment. However, such an approach could lead to a demotion to a one-level policy. Another way is to constantly monitor the dynamics of the environment during the sub-goal pursuit, and if circumstances so require, be ready to change the designated sub-goal. This approach seems to align with the human planning process, where, for example, driving a fixed route to work is modified when a road accident occurs.

In our proposed approach, the higher-level control constantly verifies whether the current sub-goal should not be replaced by a different one, more appropriate for the current state. In the basic scenario, the high-level agent returns the sub-goal and the time it should take to complete it. However, a high-level agent, instead of being active only every certain number of environment steps, receives an observation after each bottom-level agent step. Based on this, a decision is made whether we are still pursuing the previously set sub-goal or if we need to change it to a new one.

The contribution of this paper can be summarized as follows: 
\begin{itemize} 
\item We introduce a method, EAT, of monitoring and possibly terminating higher level actions in hierarchical RL. This method allows a hierarchical policy to immediately react to random events in the environment. 
\item We design two strategies for monitoring and terminating the higher level actions. 
\item We introduce a framework for hierarchical decomposition of Markov Decision Processes into subprocesses in which rewards for future events are discounted over time elapsing to their occurrence rather than over the number of actions to their occurrence. 
\end{itemize} 


\section{Related work} 
\label{sec:related-work} 

Hierarchical Reinforcement Learning (RL) is based on decomposing long-horizon reinforcement learning tasks into smaller sub-tasks which are chosen by the higher-level policy \citep{parr1997reinforcement,pateriaHierarchicalReinforcementLearning2021}. Sub-tasks are composed of atomic actions and might be learned with RL or hand-crafted. A sub-tasks learning, also called sub-task discovery, could be performed in parallel to learning hierarchical policy or as an independent pretraining stage of the process.

One way to introduce hierarchy into control is to decompose a~given Markov Decision Process into a~hierarchy of smaller MDPs. An action of a~higher-level MDP selects {\it an option}, which is a~lower-lever MDP which is executed until its terminal state~\cite{precup2000temporal,sutton1999between}. 
In some approaches, options are predefined~\cite{sutton1999between,barto2003recent, precup2000temporal,shankar2020learning}, i.e., separately trained or specially handcrafted.
An alternative approach is for example the Option Critic~\citep{baconOptionCriticArchitecture2016} method, in which Options (policies and subtask ending functions) are discovered from the beginning of the training of the entire hierarchical policy. One of the first papers where Option discovery was used in the problem of continuous control is~\citep{bagaria2019option}. However, an important limitation of the proposed method is the requirement that the target task includes explicit goal states.
From this family of methods, the closest approach to ours is the model proposed in~\cite{li2018learning}. The terminating function of an option is also trained, so that the set of terminal states changes with training time. The significant difference is that the test method is based on low-level deterministic options representing the knowledge of a human expert.

Another similar approach is skill-based methods~\cite{eysenbach2018diversity,sharma2019dynamics,campos2020explore}. Skills are modeled by a low-level policy conditioned on an additional variable -- different variable triggers different behavior. Usually, the~low-level agent learns skills in unsupervised way -- intrinsic reward by maximization of the mutual information (MI) between skill and trajectories resulting from using that skill. \citet{zhang2021hierarchical} introduced the HIDIO architecture, in which the~low-level agent learns skills in the unsupervised way. As part of a high-level action, the variable that triggers the appropriate skill is returned.

In another approach, the high-level policy output serves to communicate with a lower-level agent~\cite{schmidhuber1991learning,mcgovern2001automatic,vezhnevets2017feudal}. Usually, high-level information is attached to lower-level observations.
In most cases, high-level politics returns the sub-goal for low-level policies. In this scenario, the low-level agent is rewarded for approaching the designated sub-goal.  However, using previous experience to train a~hierarchy of policies raises the problem of non-stationarity. This experience taken literally is irrelevant as selecting targets for lower-level policies leads now to different results, as these policies have changed due to learning. To address this issue, different methods of subgoal re-labeling were proposed. \citet{levy2017learning} proposed hierarchical experience replay (HAC): actual states achieved are used as if they have been selected subgoals. On the other hand,~\citet{nachum2018data} presented an approach named HIRO based on \textit{off-policy correction} where the unattained subgoal is re-labeled in transition data with another one, drawn from the distribution of subgoals that maximize the probability of the observed transitions.

However, few articles have been written that consider a scenario in which low-level policy does not work for a fixed number of steps. The selection of the appropriate frequency in decision-making by a high-level policy has a significant impact on the performance of the proposed methods. 
 Too long or too short a period of time may degenerate the entire approach to a non-hierarchical method or disrupt the convergence of the algorithm.
The possibility of dynamically changing the higher-level action in hierarchical reinforcement learning was explored in \cite{zhou2020}. In the proposed TEMPLE algorithm, the temporal switch that decides if a new high-level action should be selected is part of the lower-level output.

The main difference between the above approach and our approach lies in the level at which the decision to change the subgoal is made. In the TEMPLE method, this low-level policy returns a switch signal, which determines how much, in the next step, the new high-level action will be used, and how much the present one will.

The HiTS algorithm~\cite{gurtler2021hierarchical} represents the approach opposite to TEMPLE, and still different from ours. A high-level agent not only determines the target position but also the time in which it has to be achieved. Therefore, the high-level action is a pair, $(g^0,\Delta)$, comprising a subgoal $g^0$ and the number of the low-level steps $\Delta$ in which the subgoal should be achieved.

As part of our approach, not only a low-level agent is supposed to achieve the given subgoals in a specific time, but also a high-level agent has the ability to immediately change the implemented plan and interrupt the current high-level action. The method designed in this way is better suited to operating in dynamic or random environments. Thanks to the ability to change the target, the high-level agent does not have to accurately predict the future and is able to efficiently correct the plan when, for example, it is not implemented correctly due to insufficient convergence of the low-level agent.

\section{Problem formulation} 
\label{sec:problem-formulation} 

We consider the typical RL setup \citep{2018sutton+1} based on a~Markov Decision Process (MDP): An agent operates in its environment in discrete time $t=1,2,\dots$. At time~$t$ it finds itself in a~state, $\state_t\in\stateSpace$, performs an action, $\ctrl_t\in\ctrlSpace$, receives a~reward, $r_t\in\real$, and the state changes to $\state_{t+1}$. Choosing actions, the agent anticipates future rewards with the discount factor of $\gamma\in[0,1)$. 

We assume that effective hierarchical control is possible in this MDP. Let there be $L>1$ levels of the hierarchy. Each $l$-th level defines an~MDP with its state space $\stateSpace^l$, its action space $\ctrlSpace^l$ and rewards. For the highest level $\stateSpace^L=\stateSpace$ and for the bottom level $\ctrlSpace^1=\ctrlSpace$. Taking an action at $l$-th level, $l\geq2$, $\ctrl^l_t$, launches an episode of the MDP at $l-1$-st level. $\ctrl^l_t$ defines the goal in this lower-level episode, thus the states and rewards in this episode are co-defined by $\ctrl^l_t$. Once this episode is finished, another action at $l$-th level is taken. 

Actions at the $l$-th level are defined by a~policy, 
\Beq \label{policies} 
    \ctrl^l_t = \pi^l(\state^l_t, \xi^l_t), \; 
    l=L,\dots,1, 
\Eeq 
where $\state^l_t$ is the state of the system perceived at $l$-th level of the hierarchy, at time $t$, and $\xi^l_t$ is a~random element. For $l<L$, $\state^l_t$ represents, among others, the objective of the current operation at $l$-th level, resulting from the current $l+1$-st level action. The random element $\xi^l_t$ enables exploration and learning of this policy. 

The goal is to learn the hierarchy of policies \eqref{policies} so that in each state at each hierarchy level the expected sum of future discounted rewards is maximized. 

\paragraph{Robotic example with $L=2$.} In robotic applications of RL, state $\state_t$ is usually a~vector that comprises (i) positions of joints, (ii) velocities of the joints, (iii) readouts of sensors outside joints, (iv) a~vector that defines the current tasks. Actions at the 2-nd level of the control hierarchy define target positions (and velocities) of the joints while the 1-st control level brings the joint positions (and velocities) to the given targets. Effective control of the robot with this hierarchy is possible because the robot is only able to accomplish given tasks by means of its body.

\section{Method} 
\label{sec:method}

\subsection{Hierarchy of MDPs}
\label{sec:hierarchy-of-MDPs} 

We assume that a~decomposition of the original MDP into a~hierarchy of MDPs satisfies the following conditions: 
\begin{enumerate} 
\item 
A higher-level action corresponds to an episode of the lower-level MDP. This episode may finish with a terminal state (e.g., when the lower level goal is achieved) or with a~nonterminal state (e.g., when a~predefined timeout has been reached). 
\item 
Each hierarchy level has its own discount factor, $\gamma_l\in(0,1)$, with $\gamma_L$ equal to $\gamma$ of the original MDP. The discount factor at the given level is related to how farsighted the policy at this level should be. Typically, the policy needs to be more farsighted at higher (in words tactical or strategic) levels, thus $\gamma_l<\gamma_{l+1}$. 
\item 
At each hierarchy level, a reward, $r^l_t \in \real$, is defined for each time instant of the original time, $t$, with the rewards for the highest level equal to the original MDP rewards, $r^L_t\equiv r_t$. At each level, the sum of discounted future rewards 
\Beq 
    \sum_{i\geq0} r^l_{t+i} \gamma_l^i
\Eeq
is being maximized. 
\end{enumerate} 
Note that the above assumptions are atypical. Usually \cite{gurtler2021hierarchical}, it is assumed that each hierarchy level has its own time indexing determined by subsequent actions at this level. A~single reward is then being paid at each hierarchy level for a~single action performed at this level, and further rewards are discounted at $\gamma_l$, $\gamma_l^2$, etc. However, when the policy at this level has any control over the duration of actions, it can manipulate this duration to maximize the sum of rewards, which usually contradicts the objectives of control at this level.

For instance, suppose at the level $l$ at time $t$, a~catastrophic event is expected to happen in $\Delta$ original time instances, covering $k$ actions at this hierarchy level. In our formulation, the negative reward for this event is always discounted at $\gamma_l^{\Delta}$. When the typical formulation is adopted, that reward is discounted at $\gamma_l^k$. The policy may learn to minimize this weight just by choosing short-lasting actions, thereby maximizing $k$. 

A hierarchical RL algorithm designed for the above typical formulation can usually be adjusted to our formulation. However, it may rise some technical issues, because event-specific discounting of further rewards needs to be introduced instead of just universal discounting.

\subsection{Emergency higher-level action termination} 
\label{sec:action:termination} 

The low-level actions work to realize plans imposed by the higher level actions. However, it may happen that sticking to these plans is inefficient on unanticipated changes of the environment state. Then, the current higher level plan needs to be terminated and a~new one needs to be initiated. Following this rationale, we propose to proceed according to the following principles, applicable for all hierarchy levels above the bottom one: 
\begin{enumerate} 
\item 
A proposal action is designated at each time $t$ with the random element applied earlier to designate the currently realized action. 
\item 
The proposal action and the currently realized one are compared. 
\item 
If the proposal actions appears to be better or much different (explained below) than the currently realized one, then (1) the currently realized actions at this level and below are terminated, (2) the events with this actions, rewards gathered and the following states are applied to learning, (3) new actions, with new random elements are designated at this level and below. 
\end{enumerate} 
Our proposed approach is based on running Algorithm~\ref{alg} at each time $t$. 
\begin{algorithm}[h]
\caption{EAT: Emergency action termination for hierarchical reinforcement learning} 
\label{alg} 
\begin{algorithmic}[1]
\FOR{$l=L,\dots,2$}
    \STATE 
    $\altctrl^l_t \leftarrow \pi^l(\state^l_t,\xi^l_t)$ \# action proposal with old random element
    \IF{$\altctrl^l_t$ is better/different than $\ctrl^l_t$}
        \FOR{$i=l,\dots,1$} 
            \STATE Terminate the current action at level $i$
            \STATE Use this finished event for training 
            \STATE Designate a new action at level $i$
        \ENDFOR
        \STATE Break 
    \ENDIF
\ENDFOR
\end{algorithmic} 
\end{algorithm}

We propose two strategies to determine if the future rewards could increase thanks to the termination of the current action (the condition in Line 3 of the algorithm). Discussing these strategies, we will use the following notation. We assume that for each level $l\geq2$ an action-value function approximation of the current policy is available
\Beq
    \bar Q^l : \stateSpace^l \times \ctrlSpace^l \rightarrow \real. 
\Eeq
An action currently realized at $l$-th level is denoted by $\ctrl^l_t$. It is defined by an~action, $\ctrl^l_\tau$, selected at time $\tau\leq t$. Not necessarily $\ctrl^l_t = \ctrl^l_\tau$. For instance, the action $\ctrl^l_\tau$ may have indicated a~target point $x$ to be achieved in $\Delta$ time-steps; then $\ctrl^l_t$ defines the same target $x$ to be achieved in $\Delta+\tau-t$ time-steps. Following \eqref{policies}, we assume $\ctrl^l_\tau = \pi^l(\state^l_\tau,\xi^l_\tau)$ and $\xi^l_t\mathrel{\overset{\makebox[0pt]{\mbox{\normalfont\tiny\sffamily def}}}{=}}\xi^l_\tau$. Let a~currently proposed alternative to the on-going action $\ctrl^l_t$ be 
\Beq
    \altctrl^l_t = \pi^l(\state^l_t, \xi^l_t). 
\Eeq 

\subsubsection{Changing $Q$} \label{sec:changingQ}
In this strategy, we terminate the current action $\ctrl^l_t$ if it seems to be worse that the proposed alternative $\altctrl^l_t$. This is found when 
\Beq 
    \bar Q^l(\state^l_t, \altctrl^l_t) - \bar Q^l(\state^l_t, \ctrl^l_t)
\Eeq 
is greater than a~certain threshold related to variability of the of $\bar Q$ values. We assume this threshold equal to $\alpha \delta_t$, where $\alpha\in[0,1]$ is a~parameter and $\delta_t$ is a~recursive estimate of the standard deviation of the values $\bar Q^l(\state^l_\tau,\ctrl^l_\tau)$ for different time $\tau$ when actions at the level $l$ were started. 

\subsubsection{Changing target}
\label{sec:changingtarget}
In this strategy, we terminate the current action $\ctrl^l_t$ if it significantly differs from the proposed one $\altctrl^l_t$. To be able to determine how different are the actions from one another, we need further assumptions about the nature of the actions. Suppose there are two functions, $x:\ctrlSpace^l\bigcup\stateSpace^l \rightarrow \real^n$ and $\Delta:\ctrlSpace^l\rightarrow\real_+$, that give the following meaning to actions: $x(\ctrl^l_t)$ is a~point to which the~projection $x(\state^l_t)$ of the state $\state^l_t$ is to converge in time $\Delta(\ctrl^l_t)$. Actions are considered different, when they assume a notably different change of the state projection in comparison to their average length. The current action is terminated when 
\Beq 
    \begin{split} 
    & \left\|\frac{x(\altctrl^l_t)-x(\state^l_t)}{\Delta(\altctrl^l_t)}-\frac{x(\ctrl^l_t)-x(\state^l_t)}{\Delta(\ctrl^l_t)}\right\|
    \\ 
    & > \alpha \frac{1}{2}
    \left(\frac{\|x(\altctrl^l_t)-x(\state^l_t)\|}{\Delta(\altctrl^l_t)}+\frac{\|x(\ctrl^l_t)-x(\state^l_t)\|}{\Delta(\ctrl^l_t)}\right) 
    \end{split} 
\Eeq 
for $\alpha\in[0,2]$ being a threshold parameter. 
\subsection{EAT with HiTS} 
\label{sec:EAT+HiTS} 

The main algorithm of hierarchical RL to which we apply our proposed method of terminating actions is HiTS \cite{gurtler2021hierarchical}. In this method, higher level actions define goals for state projections along with time to achieve these goals, as discussed in Section~\ref{sec:changingQ}. At each hierarchy level a~plain RL algorithm is used to learn $\pi^l$ such as SAC \cite{2018haarnoja+3}. State-of-the-art efficiency of HiTS results primarily from the use of hindsight action relabeling: Suppose at a~certain hierarchy level the target state projection is missed, and another final state projection is reached instead. Then, the whole episode of reaching this final projection is added to the experience for this learning level as if this final projection was the actual target. 

Introducing emergency action termination into HiTS requires the following changes: 
\begin{enumerate} 
\item 
Rewards at each level now needs to be defined for each instant of the original MDP. 
\item 
The discounting needs to be changed as discussed in Section~\ref{sec:hierarchy-of-MDPs}. (That  entails adjustments in the plain RL algorithms that operate at each level of the hierarchy.) 
\item 
Introducing Algorithm~\ref{alg} (the conditional action termination) at each time-step~$t$. 
\end{enumerate} 

\section{Experimental study}
\label{sec:experiments} 

In this study, we evaluate experimentally the approach to hierarchical RL introduced in this paper. We compare four algorithms: 
\begin{itemize} 
\item The state-of-the-art HiTS algorithm \cite{gurtler2021hierarchical}. 
\item EAT(Q) --- the algorithm presented in Section~\ref{sec:EAT+HiTS} with the action termination strategy of Section~\ref{sec:changingQ}. 
\item EAT(geom) --- the algorithm presented in Section~\ref{sec:EAT+HiTS} with the action termination strategy of Section~\ref{sec:changingtarget}. 
\item HiTS+VariableDiscountSAC --- for an ablation, we also analyze the behavior of HiTS with the modified discounting scheme presented in Section~\ref{sec:hierarchy-of-MDPs}. This scheme is a~prerequisite to EAT but it actually is operational alone. 
\end{itemize} 

We perform experiments on seven benchmark environments. Five of them are taken from\cite{gurtler2021hierarchical}: Pendulum, Platforms, Drawbridge, UR5Reacher, Ant4Rooms. We also present results on two new environments that build on Drawbridge and Platforms but include stochastic elements. They are:
\begin{itemize}
    \item \textbf{NoisyDrawbridge} -- similar to Drawbridge environment, but there are three different times of drawbridge openings: 400, 500 or 600. Thus, in every episode the drawbridge is opened at different time step, which prevents the agent from replaying the same sequence of actions in every episode. 
    
    \item \textbf{NoisyPlatforms} -- built on Platforms environment, but with an element of uncertainty: One or both platforms can be immediately frozen with a chosen probability at every time step. 
    We freeze the active platform with probability 0.005 at every time step. Moreover, we set the maximum number of freezes of this platform to two.
\end{itemize}



Experimental settings are borrowed from \citet{gurtler2021hierarchical}. For the EAT algorithm, we use the threshold parameter (Section \ref{sec:action:termination}) $\alpha=0.5$ for EAT(Q) and $\alpha=1$ for EAT(geom) in all the experiments.

\subsection{Results}
\label{sec:exp_hits_eat} 

\begin{figure*}[h!]
    \begin{tabular}{c c}
    \includegraphics[width=0.48\textwidth]{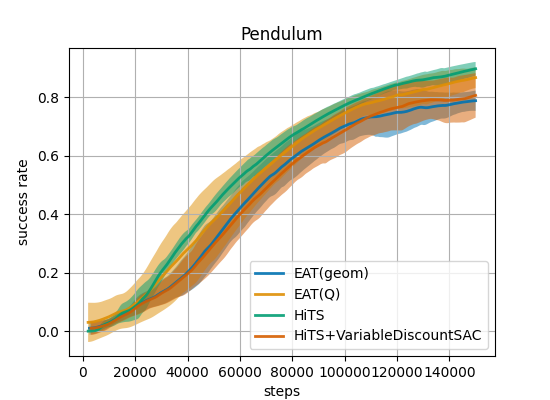} &\!\!\!\!\!
    \includegraphics[width=0.48\textwidth]{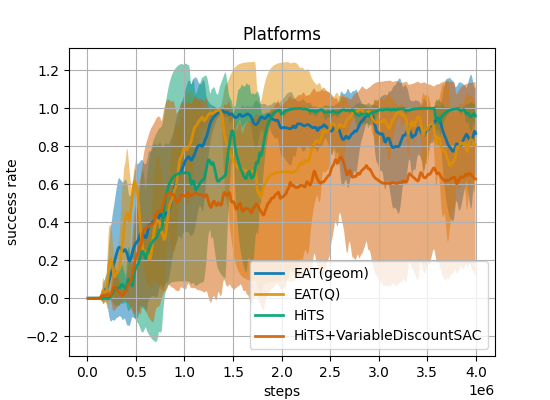} \\
    \includegraphics[width=0.48\textwidth]{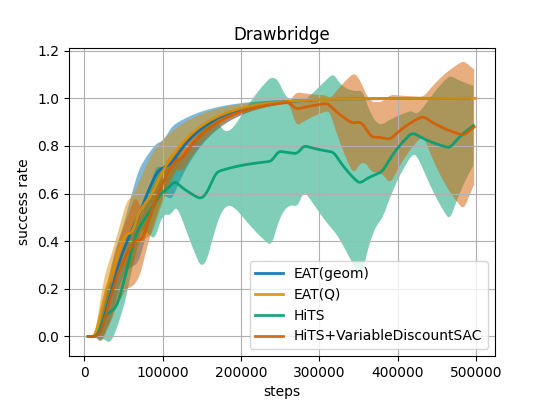} &\!\!\!\!\!
    \includegraphics[width=0.48\textwidth]{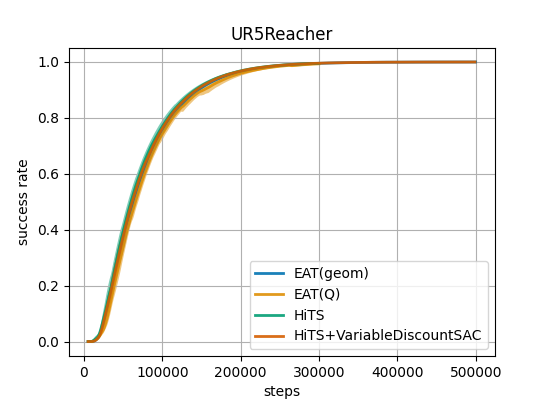}
    \end{tabular}
    \caption{Success rate in training time for  Pendulum, Platforms, Drawbridge and UR5Reacher.}
    \label{fig:results}
\end{figure*}

The results of the qualitative comparison of HiTS, EAT(Q), EAT(geom), and HiTS+VariableDiscountSAC are presented in Figures~\ref{fig:results}. We find, that terminating high-level actions is beneficial even in deterministic environments such as Drawbridge and makes no difference in terms of performance in others. EAT(Q) and EAT(geom) interruption mechanisms provide means to excel in this environment and achieve nearly perfect success rate. Meanwhile, the introduction of VariableDiscountSAC merely improves the time of learning in Drawbridge and does little change in other environments.   

Qualitatively, we find that both EATs versions perform considerably better in NoisyDrawbridge and NoisyPlatforms environments. Sudden changes in environment elements cannot be dealt with well in HiTS, due to unchangeable goals for lower-level politics. As expected, EATs learn to interrupt the invalid subgoals closely after the first indication of the current subgoal lapse. This phenomenon is studied in depth in Section~\ref{sec:eat_analysis}.

\begin{figure*}[h!]
    \begin{tabular}{c c}
    \includegraphics[width=0.48\textwidth]{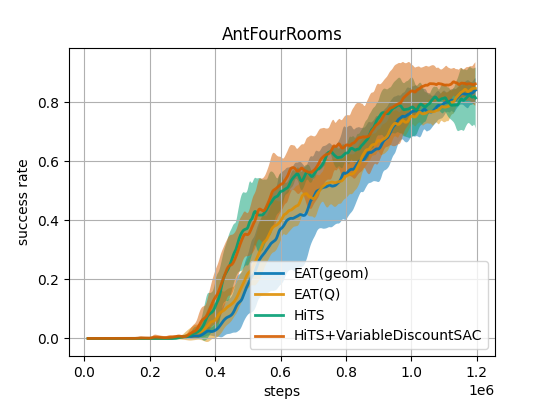} &\!\!\!\!\!
    \includegraphics[width=0.48\textwidth]{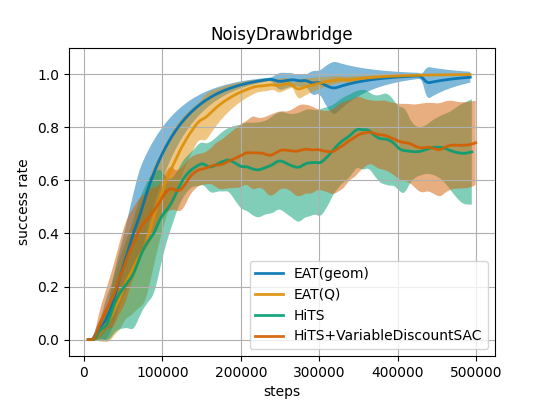} \\
    \includegraphics[width=0.48\textwidth]{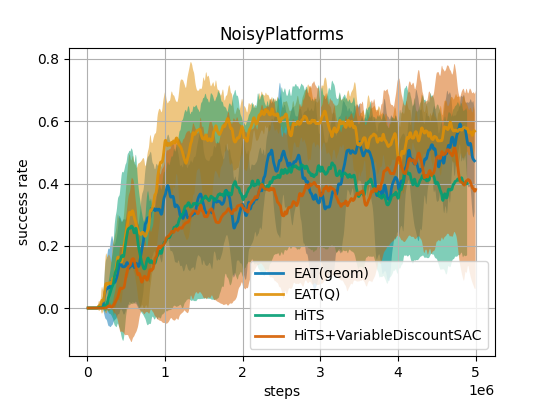} &\!\!\!\!\!
    \end{tabular}
    \caption{Success rate in training time for Ant4Rooms, NoisyDrawbridge and NoisyPlatforms.}
    \label{fig:other:results}
\end{figure*}

\subsection{Action termination analysis}
\label{sec:eat_analysis} 

\begin{figure*}
    \begin{tabular}{c c}
    \includegraphics[width=0.48\textwidth]{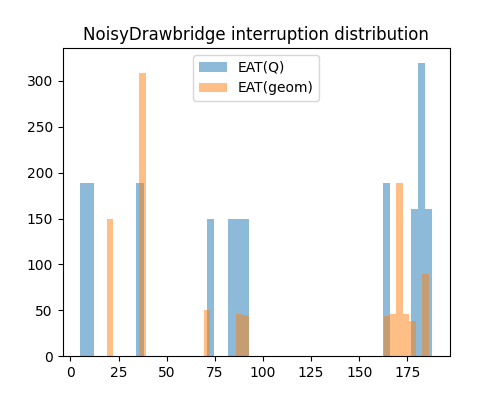}
    &
    \includegraphics[width=0.48\textwidth]{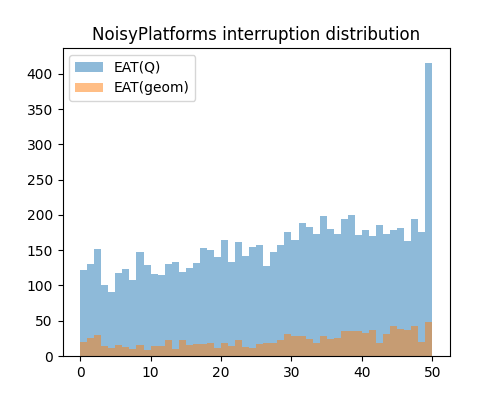}
    \end{tabular}
    \caption{Emergency action termination - interruptions relative to random events in the NoisyDrawbridge and NoisyPlatforms environments. The time steps in plots are limited to a maximal high-level action length.}
    \label{fig:interrupts}
\end{figure*}

We analyzed the behavior of models trained in experiments described in \ref{sec:exp_hits_eat} to verify if the emergency action termination mechanism works as intended by responding to random changes in an environment. We used 5 models trained using EAT(Q) and EAT(geom) methods on NoisyDrawbridge and NoisyPlatforms environments. Over the course of 500 episodes for each model, we measured the delay between the occurrence of a random event and the following interruption. The distribution of these delays is presented in the figure \ref{fig:interrupts}.

It is seen that there is a large number of interruptions in the time steps following the unexpected change in the environment. This is best seen in the NoisyDrawbridge environment, where many actions are changed within the first 100 steps after the drawbridge starts to open. It should be noted that the interruptions in this environment do not happen immediately after the random event, as it takes precisely 63 steps until the bridge allows the ship to sail through. It should also be noted, that the interruption times concerning the environment change are similar for models trained using both EAT variants. Apparently, this behaviour is the cause of the superior performance of EAT methods on the NoisyDrawbridge environment, as seen in the fig.~\ref{fig:other:results}.

The randomness of the NoisyPlatforms environment has no such strict relation with changing the possible options for the agent as in the NoisyDrawbridge environment. This is most likely caused by the fact that freezing a platform alone is not necessarily a~reason to change the high-level action immediately. However, it changes the trajectory of events in such a~way that a~better action may be selected after some time. This is reflected by a much smoother distribution of interruptions following the random events in this environment.

\subsection{Discussion} 
\label{sec:discussion}

Our experiments (HiTS+VariableDiscountSAC) show that only changing the reward and discounting pattern from that over high-level actions to that over low-level time does not affect the performance consistently. However, it is an essential prerequisite for terminating higher-level actions.  

In four out of seven our analyzed environments (Drawbridge, AntFourRooms, NoisyDrawbridge, NoisyPlatforms) EAT achieved a~higher success rate than HiTS. In one case (UR5Reacher) both algorithms reached the same success rate. Note that the original environments are deterministic, nothing unexpected happens there, and the only reason to terminate an~action is the unexpectedly low quality of this action. However, in the environments with more randomness, NoisyDrawbridge and NoisyPlatforms, there are objective reasons to terminate actions and EAT consequently produces significantly higher ultimate success rates than HiTS. 
\section{Conclusions} 
\label{sec:conclusions} 

In this paper, we introduced a framework for dividing a Markov Decision Process into subprocesses in which rewarding and discounting are over the original MDP time. In this framework, we introduced a~method of terminating higher-level actions when they become obsolete due to random events in the environment. Our proposed method enables an immediate response of control to these events, thereby increasing control quality. The experiments confirm that the quality increases indeed, especially in non-deterministic environments.

\wyciete{
\section*{Acknowledgments}

This work was partially funded by a grant from Warsaw University of Technology Scientific Discipline Council for Computer Science and Telecommunications.
}

\bibliographystyle{ACM-Reference-Format}
\ifdefined\headicml
\bibliographystyle{icml2022}
\fi 
\ifdefined\headijcai
\bibliographystyle{named}
\fi
\ifdefined\headneurips
\bibliographystyle{apalike} 
\fi
\ifdefined\headaamas
\renewcommand{\emph}{\textit}
\bibliographystyle{ACM-Reference-Format}
\fi

\bibliography{references}

\appendix
\input{appendix}

\end{document}

%% file: macros.tex
\def\state{s}
\def\stateSpace{\mathcal{S}}
\def\ctrl{a}
\def\ctrlSpace{\mathcal{A}}
\def\altctrl{\mathring a}

\def\real{\mathbb R}

\def\comment#1{}

\def\eqref#1{(\ref{#1})}
\def\Beq#1\Eeq{\begin{equation}#1\end{equation}}
\def\Beqo#1\Eeqo{\begin{equation*}#1\end{equation*}}
\def\Beqs#1\Eeqs{\begin{align}#1\end{align}}
\def\Beqso#1\Eeqso{\begin{align*}#1\end{align*}}

%% file: appendix.tex
\section{Algorithms' hyperparameters} 
\label{AlgorithmsHyperparams}

Hyperparameter of HiTS have been taken from \cite{gurtler2021hierarchical}. Common parameters for all algorithms are presented in Tab.~\ref{tab:params:platform_drawbridge} for Platforms and Drawbridge environments and in Tab.~\ref{tab:params:pendulum_reacher_ant} for Pendulum, UR5Reacher and AntFourRooms environments. Hyperparameters specific for HiTS algorithm with original SAC and VariableDiscountSAC are presented in Tab.~\ref{tab:params:HiTS} and Tab.~\ref{tab:params:VarDiscount} respectively. Hyperparameters of EAT(Q) and EAT(geom) are presented in Tab.~\ref{tab:params:EATQ} and Tab.~\ref{tab:params:EATgeom} respectively. Noisy environments introduced in \ref{sec:experiments} share hyperparameters with their original versions.

\begin{table*}
    \centering
    \begin{tabular}{c|c|c}
        \hline
        Parameter & Platforms (+NoisyPlatforms) & Drawbridge (+NoisyDrawbridge)\\
        \hline
        \multicolumn{1}{l|}{High-level}&\\
        Models size & $\langle32,32\rangle$ & $\langle32,32\rangle$ \\
        Learning rate & $1.940204674106782\cdot10^{-4}$ & $7.227658105394519\cdot10^{-5}$ \\
        Discount & 0.99 & 0.99 \\
        Flat algorithm & SAC & SAC \\ 
        SAC $\alpha$ & $4.189083521541997\cdot10^{-3}$ & $2.0709754482693517\cdot10^{-2}$ \\
        SAC $\tau$ & $2.1421017364015173\cdot10^{-2}$ & $3.143822236379807\cdot10^{-1}$ \\
        Hindsight goals & 3 &3 \\
        Batch size & 512 & 256\\
        Random action fraction & 0.05 &0.05 \\
        Learning start & 20000 & 0 \\
        Grad steps/env steps & 0.5 & 1 \\
        Max actions in episode & 10 & 5\\
        \hline
        \multicolumn{1}{l|}{Low-level}&\\
        Models size & $\langle32,32\rangle$ & $\langle32,32\rangle$ \\
        Learning rate & $1.940204674106782\cdot10^{-4}$ & $7.227658105394519\cdot10^{-5}$ \\
        SAC $\alpha$ & 1.1172711243974338 & $5.413320369694484\cdot10^{-2}$\\
        SAC $\tau$ & $2.1421017364015173\cdot10^{-2}$ & $3.143822236379807\cdot10^{-1}$\\
        Hindsight goals & 3 &3 \\
        Batch size & 512 & 256\\
        Random action fraction & 0.05 &0.05 \\
        Deterministic transitions & 0.3 & 0.3\\
        Learning start & 20000 & 0 \\
        Grad steps/env steps & 0.5 & 1 \\
        \hline
    \end{tabular}
    \caption{Common hyperparameters for Platforms, Drawbridge, and noisy variants of these environmnets}
    \label{tab:params:platform_drawbridge}
\end{table*}

\begin{table*}
    \centering
    \begin{tabular}{c|c|c|c}
        \hline
        Parameter & Pendulum & UR5Reacher & AntFourRooms\\
        \hline
        \multicolumn{1}{l|}{High-level}&&\\
        Models size & $\langle32,32\rangle$ & $\langle64,64,64\rangle$& $\langle64,64,64\rangle$ \\
        Learning rate & $6.441137873509102 \cdot10^{-3}$ &$4.968380769591525\cdot10^{-4}$ & $2.732962767235062\cdot10^{-4}$ \\
        Discount & 0.99 & 0.99 & 0.99 \\
        Flat algorithm & SAC & SAC & SAC \\ 
        SAC $\alpha$ & 2.525263906546272 && 1.1266973662447297 \\
        SAC target entropy & & -13.223820062136808 &\\
        SAC $\alpha$ optimization & & 0.001 & \\
        SAC $\tau$ & $1.258608875021\cdot10^{-2}$ &$1.1987881761644309\cdot10^{-2}$& $17099210322824967\cdot10^{-3}$ \\
        Hindsight goals & 3 &3&3 \\
        Batch size & 256 &1024&1024 \\
        Random action fraction & 0.05 & 0.05 & 0.05 \\
        Learning start & 0 &0& 0\\
        Grad steps/env steps & 1 &0.07& 0.07 \\
        Max actions in episode & 22 & 24 & 22 \\
        \hline
        \multicolumn{1}{l|}{Low-level}&&\\
        Models size & $\langle32,32\rangle$ & $\langle64,64,64\rangle$& $\langle64,64,64\rangle$ \\
        Learning rate & $6.441137873509102 \cdot10^{-3}$ &$4.968380769591525\cdot10^{-4}$& $9.288851022175569\cdot10^{-4}$ \\
        SAC $\alpha$ & &$5.317222522776922\cdot10^{-4}$& $2.609605666772228\cdot10^{-3}$\\
        SAC target entropy & -2.60162130869488 & &\\
        SAC $\alpha$ optimization & $6.441137873509102 \cdot10^{-3}$ & & \\
        SAC $\tau$ & $1.258608875021\cdot10^{-2}$ &$1.1987881761644309\cdot10^{-2}$& $17099210322824967\cdot10^{-3}$ \\
        Hindsight goals & 6 &3&3 \\
        Batch size & 256 &1024&1024 \\
        Random action fraction & 0.05 &0.05& 0.05\\
        Deterministic transitions & 0.3 &0.3& 0.3\\
        Learning start & 0 &0&0 \\
        Grad steps/env steps & 1 &0.07& 0.07\\
        \hline
    \end{tabular}
    \caption{Common hyperparameters for Pendulum, UR5Reachet, and AntFourRooms}
    \label{tab:params:pendulum_reacher_ant}
\end{table*}

\begin{table*}
    \centering
    \begin{tabular}{c|c}
    \hline
    Parameter & Value \\
    \hline
    HL flat algorithm & SAC \\
    \multicolumn{1}{l|}{Platforms(+Noisy)}\\
    HL Discount & 0.97\\
    \multicolumn{1}{l|}{Drawbridge(+Noisy)}\\
    HL Discount & 0.97 \\
    \multicolumn{1}{l|}{Pendulum}\\
    HL Discount & 0.99 \\
    \multicolumn{1}{l|}{UR5Reacher}\\
    HL Discount & 0.97 \\
    \multicolumn{1}{l|}{AntFourRooms}\\
    HL Discount & 0.97 \\
    \hline
    \end{tabular}
    \caption{HiTS-specific hyperparameters.}
    \label{tab:params:HiTS}
\end{table*}

\begin{table*}
    \centering
    \begin{tabular}{c|c}
    \hline
    Parameter & Value \\
    \hline
    HL flat algorithm & VarDiscountSAC \\
    HL Discount & 0.999 \\
    \hline
    \end{tabular}
    \caption{HiTS+VarDiscountSAC hyperparameters.}
    \label{tab:params:VarDiscount}
\end{table*}

\begin{table*}
    \centering
    \begin{tabular}{c|c}
    \hline
    Parameter & Value \\
    \hline
    HL flat algorithm & VarDiscountSAC \\
    HL Discount & 0.999 \\
    $\alpha$ & 0.5 \\
    Exponential average of Q smoothing& 0.999\\
    \hline
    \end{tabular}
    \caption{EAT(Q) hyperparameters.}
    \label{tab:params:EATQ}
\end{table*}

\begin{table*}
    \centering
    \begin{tabular}{c|c}
    \hline
    Parameter & Value \\
    \hline
    HL flat algorithm & VarDiscountSAC \\
    HL Discount & 0.999 \\
    $\alpha$ & 1 \\
    \hline
    \end{tabular}
    \caption{EAT(geom) hyperparameters.}
    \label{tab:params:EATgeom}
\end{table*}